\documentclass[conference]{IEEEtran}
\IEEEoverridecommandlockouts
\usepackage{amsmath,amssymb,amsfonts}
\usepackage{graphicx}
\usepackage{textcomp}
\usepackage{xcolor}
\usepackage{algorithm}
\usepackage{algpseudocode}
\usepackage{subcaption}
\usepackage[hidelinks]{hyperref}
\usepackage{booktabs}
\def\BibTeX{{\rm B\kern-.05em{\sc i\kern-.025em b}\kern-.08em
    T\kern-.1667em\lower.7ex\hbox{E}\kern-.125emX}}

\begin{document}

\title{
Understanding Patterns of Deep Learning Model Evolution in Network Architecture Search
\thanks{Funding was provided by US Department of Energy}
}

\author{\IEEEauthorblockN{Robert Underwood}
\IEEEauthorblockA{\textit{MCS Division} \\
\textit{Argonne National Laboratory}\\
Lemont, United States \\
runderwood@anl.gov}
\and
\IEEEauthorblockN{Meghana Madhastha}
\IEEEauthorblockA{\textit{Computer Science} \\
\textit{Johns Hopkins University}\\
Baltimore, United States\\
mmadhya1@jhu.edu}
\and
\IEEEauthorblockN{Randal Burns}
\IEEEauthorblockA{\textit{Computer Science} \\
\textit{Johns Hopkins University}\\
Baltimore, United States \\
randal@cs.jhu.edu}
\and
\IEEEauthorblockN{Bogdan Nicolae}
\IEEEauthorblockA{\textit{MCS Division} \\
\textit{Argonne National Laboratory}\\
Lemont, United States \\
bnicolae@anl.gov}
}

\maketitle

\begin{abstract}
Network Architecture Search and specifically Regularized Evolution is a common way to refine the structure of a deep learning model.
However, little is known about how models empirically evolve over time which has design implications for designing caching policies, refining the search algorithm for particular applications, and other important use cases.
In this work, we algorithmically analyze and quantitatively characterize the patterns of model evolution for a set of models from the Candle project and the Nasbench-201 search space.
We show how the evolution of the model structure is influenced by the regularized evolution algorithm.
We describe how evolutionary patterns appear in distributed settings and opportunities for caching and improved scheduling.
Lastly, we describe the conditions that affect when particular model architectures rise and fall in popularity based on their frequency of acting as a donor in a sliding window.
\end{abstract}

\begin{IEEEkeywords}
Transfer Learning, AI, Network Architecture Search, Regularized Evolution, Characterization Study
\end{IEEEkeywords}
\section{Introduction}
Network Architecture Search (NAS) is a foundational method for identifying viable deep learning (DL) model
architectures suitable to solve a variety of problems. Unlike trail-and-error approaches that are time-consuming
and whose quality depends heavily on the experience of DL experts, NAS explores a large number of candidate 
models from a search space that is based on a set of rules that define what choices are possible and how they can be combined to obtain valid DL model candidates. With the increasing complexity of the problems solved with DL,
NAS is quickly becoming the only viable approach to producing high-quality DL model architectures. It has been used for Cancer Research \cite{wozniak_candlesupervisor_2018}, to shape the development of large foundation models \cite{nvidia_megatron-lm_2023}, and to design models that optimize the performance of particular applications \cite{egele_agebo-tabular_2021}. 

NAS is a computationally and resource-intensive process that often takes hundreds of state-of-art GPUs to 
find good candidates. It is difficult for two reasons: 1) the search space of DL model architectures is huge --  
in some cases larger than $10^{57}$ possible candidates, which are far too large to exhaustively search even when distributed over an entire HPC system, and 2) evaluating each candidate is very expensive, taking in some cases multiple minutes to hours to train even a single epoch using state of the art GPUs. Therefore, the problem
of how to scale NAS is critical.

Frameworks such as DeepHyper \cite{balaprakash_autotuning_2018,egele_agebo-tabular_2021} adopt
a master-worker paradigm to scale NAS on supercomputing resources. Specifically, the master is responsible
to generate new DL model candidates and pass them to workers for evaluation. However, a naive strategy that randomly
samples the search space to generate new DL model candidates performs poorly, generating DL model candidates
of low quality for a given amount of time and resources spent. 

As a consequence, more informed strategies have been proposed to generate DL model candidates. One
such popular strategy is \emph{regularized evolution}~\cite{real_regularized_2019}, which is inspired by
genetic algorithms. It consists of two stages: 1) generate an initial random population of $N$ candidates; 2) evolve the population of $N$ candidates by randomly selecting a subset $K \in N$, take the best performer in $K$,  3) perform a single mutation of the architecture of the best performer to produce a new candidate, 
4) train the new candidate to obtain a score; 5) replace the oldest model in the population with the new result. 

Regularized evolution can be complemented by \emph{transfer learning}~\cite{tan_survey_2018} to further
increase the quality of the identified DL model candidates and to reduce the duration of the search
~\cite{liu_accelerating_2021}. Specifically, given two related problems A and B, if we already have a trained DL model $M_A$ to solve A, then, instead of training a new DL model $M_B$ from scratch to solve B, we could start from a variation of $M_A$ that retains some or all of the layers in $M_A$ while initializing the new layers with random weights. In this case, we provide a better starting point for $M_B$ that ``transfers'' the knowledge from $M_A$, which is likely to make $M_B$ converge faster. In addition, the transferred layers in $M_B$ are often ``frozen'' during the training, which means they are not updated during the back-propagation, thereby accelerating every training iteration. At the same time, the amount of data needed to perform the training can often be reduced without sacrificing accuracy. These aspects combined lead to a significant reduction in training duration~\cite{chollet_deep_2018}, assuming there is an efficient DL model repository that hides the 
I/O overheads of storing/loading the weights of the DL models necessary to perform the 
transfer learning in a scalable manner~\cite{madhyasthaDStoreLightweightScalable2023}.

However, techniques to accelerate NAS with transfer learning and to produce DL models of better quality
requires a deep understanding of how the candidates are generated and evolved.
In our paper, we attempt to characterize the candidates that are identified during NAS under regularized evolution to enable future improvements to systems software conducting NAS by answering the following research questions through a combination of empirical study and algorithmic analysis:
(RQ1) How does the architecture of the candidates evolve structurally (e.g., do popular mutations appear  earlier or later in the candidate architecture) and does it change over time? 
(RQ2) How do model evolution patterns change in the context of asynchronous distributed workers where there is incomplete knowledge of model performance?
(RQ3) When does a candidate become popular (and therefore a frequent source of transfer learning) and when does it become less popular (and therefore less relevant)?
(RQ4) How does the DL model candidate quality evolve over time during NAS?

Answers to these questions have broad implications for the design of scalable NASs including caching strategies of popular tensors used for transfer learning by the search strategy (both regularized evolution and other genetic algorithms), distribution strategy of candidates to workers for evaluation, and improvements to the search strategy itself. Specifically,
by understanding when models become popular, we can design efficient caching and prefetching techniques targeted at groups of tensors used during transfer learning that leverage probabilities that a candidate will actually become a popular transfer donor. By understanding where mutations occur in the graph of a candidate architecture, we can understand better how many layers might need to be retrained after transfer or differ substantially from the architecture of the previous candidates, which we can use to refine the search strategy. Additionally, if top-performing candidates tend to have mutations in the layers close to the end of the architecture, we can bias the search strategy to favor such mutations and need to retrain fewer layers.
By understanding when a model falls out of popularity, we can design cache eviction strategies to best utilize effective cache space. Furthermore, answering these questions would allow the refinement of heuristics to optimize transfer learning in the context of NAS.

This paper contributes with a case study that involves real-world DL models from the Candle Project~\cite{wozniak_candlesupervisor_2018} and test problems from NAS benchmarks~\cite{naslib-2020} to bring qualitative and quantitative answers to these questions.
We summarize our key contributions as follows:

\begin{itemize}
    %Note: prefix is used without being defined
    \item We introduce a characterization methodology for the patterns emerging in the evolution
    of the DL model candidates during NAS. In particular, we insist on general aspects applicable to
    all research questions, such as: 1) selecting search spaces; 2) how to efficiently evaluate the search
    spaces; 3) how to instrument the DL tools used for large-scale NAS and how and efficiently collect comprehensive traces without compromising the performance (Section~\ref{sec:characterize}). 
    
    \item We study how the architecture of DL model candidates evolves structurally over time. In doing so
    we aim to answer the question of whether mutations tend to appear earlier or later in the model architecture over time. We show that random selection of both location and specific mutations has an impact on the kinds of models observed during NAS (Section~\ref{sec:structure})
    
    \item We study how model evolution patterns occur in a distributed setting. In particular, we
    show that 1) there are temporal localities of accessing particular tensor both with respect to
    a single worker and across workers; 2) we can leverage these temporal localities by delay scheduling
    decisions for the purpose of co-scheduling groups of candidate evaluations that share a common structure
    (Section~\ref{sec:distributed})

    \item We study how the popularity of DL model candidates evolves during NAS and what conditions trigger
    the rise and fall in popularity. In particular, we show that models can be classified into popularity
    tiers and that we can determine thresholds for when a model moves between popularity tiers based on
    the frequency of it acting as a donor within a sliding window. (Sections~\ref{sec:time} and~\ref{sec:popularity}).
\end{itemize}

In the subsequent sections, we describe the NAS process, summarize prior characterization studies for genetic algorithms, describe our experimental methodology, present the results of our study, and discuss them in the context of our research questions.

\section{Background}

While deep neural networks have been tremendously successful at learning tasks. However, existing state-of-the-art networks have been manually designed by researchers. Neural Architecture Search enables us to automate architecture engineering by searching through a series of potential architectures, estimating the performance of each architecture on a dataset and finding the best one. A typical neural architecture workflow consists of the following steps. First, a search space is constructed. The search space is a graph with each edge(in NASLib\cite{naslib-2020}) or node(in DeepHyper\cite{balaprakash_DeepHyper_2018}) consisting of a number of operations that we can choose from thus creating a combinatorial search space. The graph itself is fixed beforehand with fixed blocks/layers from prior wisdom of building architectures. The next component is the search algorithm itself. Search spaces consist of millions of potential architectures to choose from. The search algorithm determines the order in which these architectures are searched. One such method is to perform a brute-force search at random. However, one can use information about models already evaluated to guide the search. Genetic algorithms and Bayesian optimization are the most commonly used state-of-the-art search techniques. In this work we will focus on regularized evolution but similar principles apply to other search methods too. The third component is performance predictions. Training each candidate architecture for a large number of epochs is computationally efficient. There exist various techniques to estimate the architecture performance without fully training it. Here, we train the model for a small number of epochs.

% Regularized Evolution is a type of genetic algorithm used in and commonly used in NAS.
% NAS is a method of optimizing a models' architectural structure by considering a search space of candidate architectures.
% Often the search space is specified as a set of variable nodes \footnote{or isomorphically edges can be variable.  DeepHyper uses variable nodes \cite{balaprakash_DeepHyper_2018}, while naslib uses variable edges \cite{}} in a directed acyclic graph.
% Each of these variable nodes can take on a concrete implementation which is often a single layer, but can be an entire submodel.
% The search algorithm is responsible for choosing the order in that candidates are considered (e.g. trained, evaluated) from the search space.
% The NAS problem is a combinatorial search problem -- that is there are non-linear effects on model quality for each choice of a variable node.
% Genetic algorithms especially Regularized Evolution and Bayesian optimization are leading search algorithms.

Genetic algorithms like regularized evolution are characterized by how the approach initialization, selection, evolution, and retirement.
Initialization is how the initial population of candidates is created -- often by random selections from the search space.
Once the population is initialized, the selection process determines a subset of models from the population to ``evolve'' via mutation.
The mutation process describes how a selected candidate undergoes changes to form a new candidate solution.
As new better performing models are explored, a retirement process determines which is removed from the population for future mutation.
We specifically use the parallel implementation of Regularized Evolution from DeepHyper\cite{balaprakash_DeepHyper_2018}
A listing of the model algorithm is provided in Algorithm~\ref{alg:regevo}.
We discuss its implementation more in Section~\ref{sec:characterize}.

%\todo[inline]{
%The following paragrahs look like related work. If you discuss them in detail, why aren't we considering them
%in this paper? Instead you should focus on how to combine regularized evolution with transfer learning
%(e.g. mention the need for a smart repository, cite ICS paper, etc.)
%}

As we evaluate and mutate architectures, we would like to save these models so they can be referenced in the future. For example, if we would like to transfer weights from an existing model in order to speed up the training process, we need the capability to save models in the repository and retrive them on demand. To determine the best model to transfer from, we match the metadata graphs and find the model with the longest common metadata\cite{liu_accelerating_2021}. Previous work \cite{madhyasthaDStoreLightweightScalable2023} looks at building a model repository designed to store and access these models efficiently.

% Bayesian optimization methods instead of mutating prior explored candidates to stocastically improve a population,  instead rely on refining a modeled ``prior'' probability distribution that estimates the model performance over the entire search space conditioned upon observations of various model candidates \cite{garnett_bayesian_2023}.  The search algorithm then chooses the candidates to evaluate that maximize the expected improvement.
% Given that Bayesian optimization is less a specific approach and more a family of approaches, a variant of this work could consider the kinds of models produced by baysian optimization methods, but given the breadth of this subfield, it would need to be narrowed to more specific algorithmic choices.

% Note: supermodels isn't an alternative approach to NAS, it is a kind of NAS
%There are a few alternative approaches to NAS.
% ``Super'' models instead of training models separately, train multiple models simultaneously by sharing model weights while training \cite{}.  For ``super'' models, a search algorithm is still needed, but it is used to select the set of layers that are ``activated'' for evaluation rather than the set of layers that are combined for training.  Thus it is still interesting to consider the models that are selected by the search algorithm.

\begin{algorithm}
\caption{Parallel Regularized Evolution}\label{alg:regevo}
\begin{algorithmic}
\Require $s \in \mathcal{N}^+, p \in \mathcal{N}^+, N \in \mathcal{N}^+, s < p < N$
\State \texttt{from random import sample}
\State \texttt{pop} $\gets$ []
\For{parallel taskloop $i = 1 \dots p$} \Comment{Stage 1}
    \State \texttt{c} $\gets$ \texttt{sample(search\_space), k=1))}
    \State \texttt{quality} $\gets$ \texttt{evalulate(c)}
    \State \texttt{lock()}
        \State $\;\;$\texttt{pop.append((c, quality))}
    \State \texttt{unlock()}
\EndFor
\For{parallel taskloop \texttt{evaluated}$= p \dots N$} \Comment{Stage 2}
    \State \texttt{lock()}
        \State $\;\;$\texttt{sampled} $\gets$ \texttt{sample(pop, k=s)}
    \State \texttt{unlock()}
    \State \texttt{best} $\gets$ \texttt{max(sampled, by=$\lambda$ i: i.quality)}
    \State \texttt{c $\gets$ mutate(best, search\_space)}
    \State \texttt{quality $\gets$ evaluate(c)}
    \State \texttt{lock()}
        \State $\;\;$\texttt{pop.pop\_left()} \Comment{discard oldest}
        \State $\;\;$\texttt{pop.append((c, quality))}
    \State \texttt{unlock()}
\EndFor
\State \Return \texttt{pop}
\end{algorithmic}
\end{algorithm}

\section{Related Work}
To the best of our knowledge, there are no papers that characterize the behavior of regularized evolution to address our specific research questions.
There are a few papers that attempt to consider the set of candidates evaluated by genetic algorithms, and none of the papers that we evaluated did this in the context of NAS and the constraints that it presents.
One such paper is \cite{whitley_evaluating_1996} looks at a set of genetic algorithms and evaluates them on their mean time to solution (quality), number of evaluations (time) on a number of problems design to be challenging (inseparable, hill climbing resilient, non-linear, non-symmetric).  While these properties are true of NAS problems, it says nothing about the kinds of candidates that appear from the search process.
The Regularized Evolution paper \cite{real_regularized_2019} does not even evaluate the algorithm in this way, instead preferring to evaluate the candidates found on their quality as measured in accuracy, the model cost in FLOPs, and time to the highest accuracy.

In contrast, our paper considers at a fine-grained level what models are selected during regularized evolution and how the model selection and accuracy improvement are related from a probabilistic perspective.

\section{Characterization Methodology}\label{sec:characterize}
First, we will discuss commonalities in the experimental setup between all of our research questions.
The goal of these research questions is to start from the structure of the models generated by the search algorithm, and then proceed to how these models evolve over the course of the search to facilitate designing systems or changes that could effect the performance of the search.
In Section~\ref{sec:results}, we will then introduce each research question in turn with the specific motivations for each question, analysis from the structure of regularized evolution that informs the answer to the questions, empirical observations, and finally takeaways for each question.

\subsection{How to Choose the Search Spaces}
As stated in the introduction, NAS is very expensive, and we are limited in the number of complete NASs that we can perform.
To complement the limited ability to perform empirical observations, we will carefully study the structure of the algorithm to inform what we expect to generalize to other NAS traces given sufficient resources to run time.
Our evaluation includes two benchmarks - one from a well-established NAS (Nasbench201 on the CIFAR-10 dataset)\cite{dong2020nasbench201}, and one from a  real-world scientific application (Candle-ATTN)\cite{wozniak_candlesupervisor_2018, egele_agebo-tabular_2021}.

Given our limited ability to conduct exhaustive evaluations, we choose our two spaces to complement each other -- Nasbench201 with CIFAR-10 has only $\approx 4.8 \times 10^5$ models, but its developers used nearly 100 TPU \footnote{Google's proprietary AI accelerator card} years to exhaustively evaluate its search space and compiled the results in a queriable database allowing us to more rapidly explore it search space and consider longer searches.  CANDLE-ATTN has a much larger search space, but
because we do not have a results database for the $\approx 3.1 \times 10^{57}$ models in the Candle ATTN search space, must utilize HPC resources in parallel in order to collect even small, but meaningful results in a reasonable amount of time.

%Note: I think this should go before the sentence - given our limited ability...
CIFAR-10 is a well-known benchmark problem is an image classification task 40k training images, 10k validation images, and 10 image classes.
The model archetype for CIFAR-10 is a convolutional network ending with a dense classifier.
CANDLE-ATTN is a cancer drug interaction model that attempts to the binary classification of a particular drug will interact with  a particular type of tumor.
Candle itself is a larger suite of benchmark problems with  ATTN representing one of the larger models both in terms of its search space, and the size of the models being searched.
Its search space is defined in \cite{egele_agebo-tabular_2021}.
The model archetype for ATTN is a dense fully connected network with skip connections.
Together these search spaces allow us to consider both large search spaces and a larger number of candidates.

\subsection{Evaluating The Search Spaces}
For each search space, we apply the parallel version of regularized evolution from DeepHyper\cite{balaprakash_DeepHyper_2018} described in Algorithm~\ref{alg:regevo}.
In the parallel setting, we make all changes to the population set atomically so that there are always $p$ candidates in the population in stage 2 when sampling of the population occurs.
In each case, we choose a total number of 1000 candidates, with a population size of 100, and a sample size of 5.
We use the quality function defined by the application -- validation accuracy in both cases.
These settings provide a consistent basis from which to study the algorithms' performance across search spaces while utilizing parallel and distributed HPC resources.

\subsection{Instrumentation of AI Tools and Collection of Traces}
As we conduct the search, we collect traces of the execution.
To conduct our experiments, we made the following modifications to DeepHyper:
1) We created a variant of the BaseTrainer class to capture at fine granularity the specific tensors that were considered during the trace process using the method from \cite{madhyasthaDStoreLightweightScalable2023} to track individual tensors.
%Note: cite hongyuan's instead of dstore
2) We implemented and integrated a primitive, greedy form of transfer learning into DeepHyper using the model repository from \cite{madhyasthaDStoreLightweightScalable2023}.
3) We integrated NASLib's CIFAR and Candle's ATTN search space into DeepHyper in order to be able to evaluate these two models
4) We modified DeepHyper's built-in trace capability to produce more detailed traces taking care to not significantly perturb the timing by avoiding additional I/O operations during the search.

These traces contain the timestamp that a model evaluation begins, the timestamp of when a model evaluation ends, the worker id that conducted the search, the architecture sequence that specifies which choices were taken for each variable node in the network architecture, the quality of the model when evaluated.
We will use these traces to analyze the search process in order to better understand the model evolution process and address our research candidates.
To make this process efficient, we gather the traces in memory on each transfer server and then write out the traces at the end of the search to avoid perturbing the runtime of the search process.
To address our research questions, we will consider different aspects of these traces to better understand the structure of searches.

We conduct our experiments on the Polaris machine at ALCF.
Polaris has hardware and software as summarized in Table~\ref{tab:hardware}.
This hardware and software is both well suited for NAS but is also representative of hardware and software at leading supercomputing centers.

\begin{table}
    \centering
    \caption{Hardware and Software}
    \label{tab:hardware}
    \begin{tabular}{lr}
    \toprule
         Hardware & Description \\
         \midrule
         CPU & 32 Core AMD EYPC Zen 3 \\
         GPU & 4 Nvidia A100 (40GB) with NVLINK (600GB/s) \\
         SSD & 2 TB \\
         RAM & 512 GB DDR4 (204.8GB/s)\\
         PCIe & 64 GB/s \\
         NETWORK & HPE Slingshot 10 \\
         \bottomrule
    \end{tabular}
\end{table}

\section{Results}\label{sec:results}
Here we will analyze the results of the traces described in Section~\ref{sec:characterize}.
In each of the following subsections, we describe how we will parse the trace files described above and how that informs the answers to each of our key research questions.
In each subsection, we will present a \textit{motivation} for asking this particular research question and motivate our choice of \textit{method} to answer the question.  After that, we will proceed to an \textit{algorithmic analysis} of the problem based on the structure of the algorithm and the insights we can obtain without running experiments.  We then present and discuss the empirical \textit{observations} that we obtain and their significance to designing systems to accelerate NAS.  Finally, we conclude each subsection with a discussion of the key \textit{takeaways} from both the algorithmic analysis and observations.

\subsection{How do model architectures evolve structurally over time -- do mutations tend to be earlier or later in the model architecture over time?}\label{sec:structure}

\begin{figure*}
    \centering
     \begin{subfigure}[b]{0.4\textwidth}
        \includegraphics[width=\columnwidth]{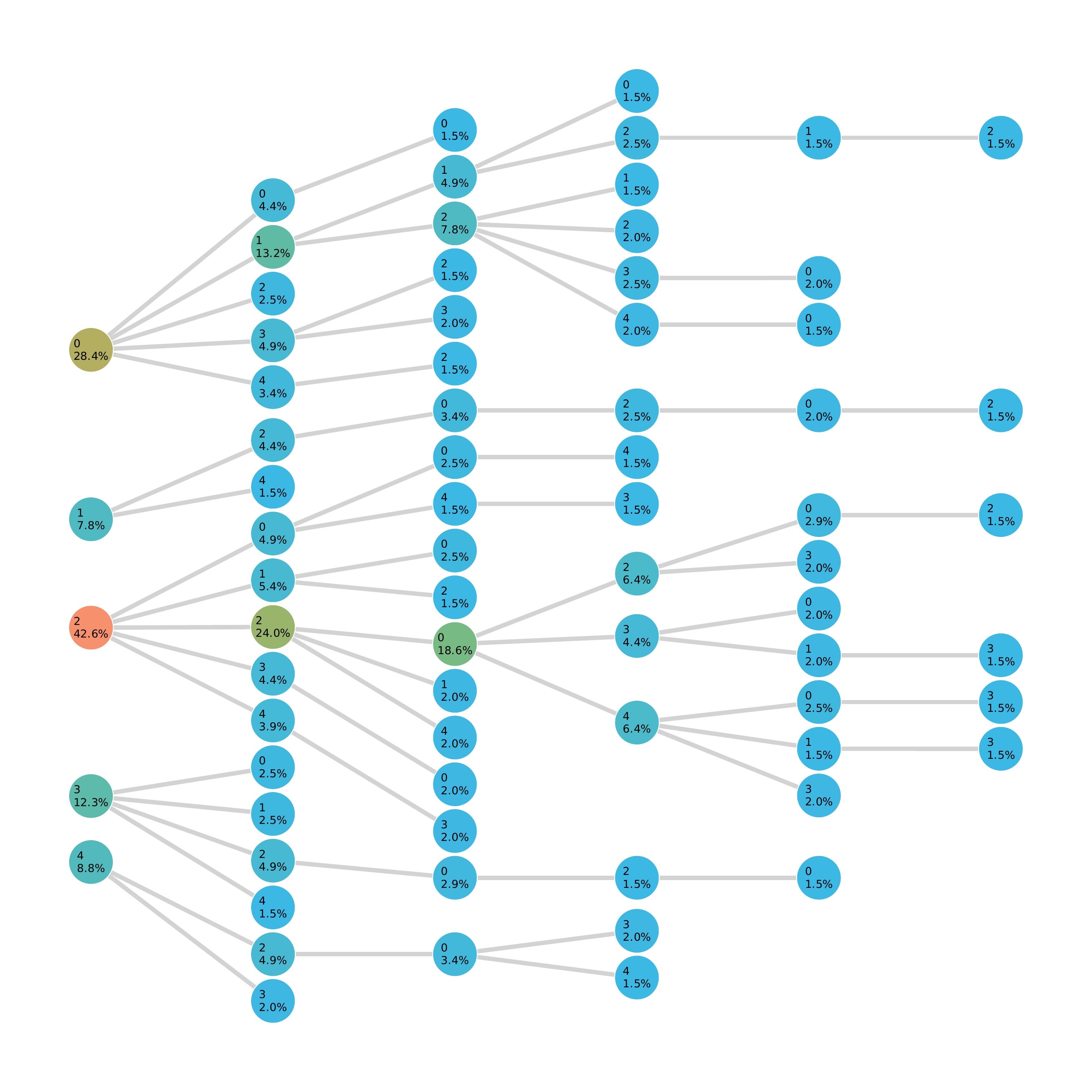}
        \caption{CIFAR - 50 epochs}
        \label{fig:triecifar50}
    \end{subfigure}%
     \begin{subfigure}[b]{0.4\textwidth}
        \includegraphics[width=\columnwidth]{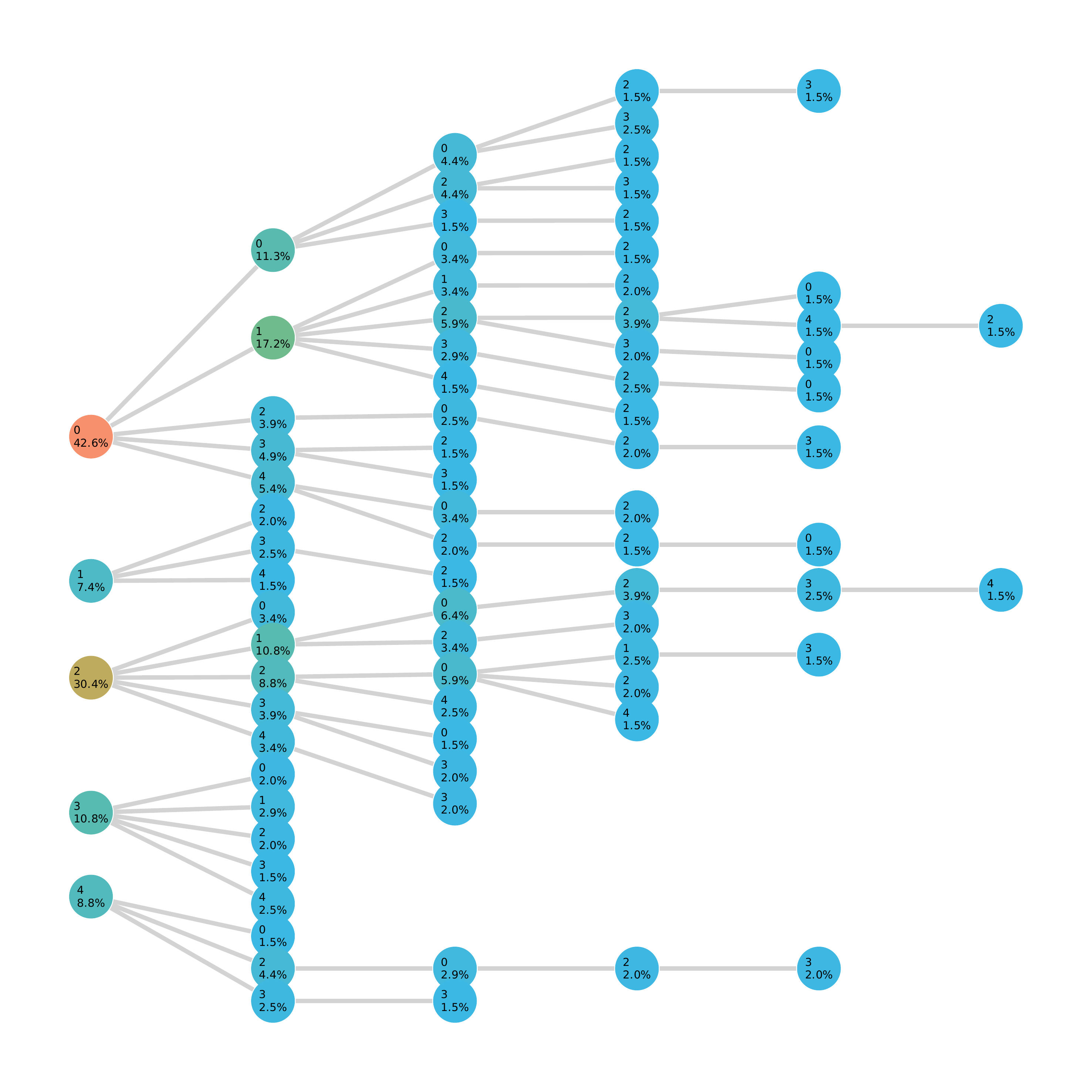}
        \caption{CIFAR - 150 epochs}
        \label{fig:triecifar150}
    \end{subfigure}
    \caption{Trie of the architecture sequences visited over the search when training for 50 and 150 epochs respectively.  Architecture sequences appearing in fewer than 1\% of the searched candidates are omitted for clarity. Color scale corresponds to the number of transfers that a model was included in.  Nodes are annotated with the variable node choice and the percentages of candidates that prefix occurred in.}
    \label{fig:trie}
\end{figure*}

\subsubsection{Motivation and Method}
We begin by studying the structural evolution of models over time.
Understanding how the structure of the models evolves over time has implications for the effectiveness of transfer learning in accelerating NAS.
For example, when transfer learning is performed, all layers that occur after a mutation must be ``invalidated'' and re-trained to account for differences in the values of their inputs.
Thus understanding where a mutation tends to occur in the structure of a model informs how often transfers occur over the course of a search process in expectation.

Tries augmented with the percentage of candidates that include a particular prefix in a search are well suited to answering the question of the overall structural evolution of the models because they concisely encode the path-dependent nature of the structure of architecture sequences visited over the course of a NAS.
As a trie, paths through the graph start on the left and proceed to the right.
During the procession, the path accumulates the various subsequences that occur over the course of the search.
For example, the top node in the rightmost column of Figure~\ref{fig:triecifar50} represents the sequence ``0-1-1-2-1-2'', and this sequence represents 1.5\% of all of the transfers that occur in the course of the search.
However, the shorter prefix of ``0-1-1'' in the same subfigure occurs 4.9\% out of all of the searches.
To enable comparisons between tries from the same application, each node is labeled with both the variable node choice ID, in addition to the percentage of architecture sequences that contain this particular variable node choice at this point in the sequence, a color code is also used where the red-er a node is the more likely it was to occur in an architecture sequence in the search, and the blue-er the less likely it was to appear.

For clarity, Figure~\ref{fig:trie} presents a subset of the variable node choices as a trie for the CIFAR search space when each model is trained for 50 or 150 epochs respectively.
Specifically, each subfigure has been pruned to remove variable node choices that appeared in fewer than 1\% of the models allowing us to focus on the most prevent prefixes which are the ones that we would like to potentially cache in transfer learning.

\subsubsection{Algorithmic Analysis}
There are two key aspects to note about the regularized evolution algorithm that affect the structure of model architecture produced by it that have implications for transfer learning and the search process overall:  1)  where regularized evolution performs its mutation and 2) how it performs its mutation after the mutation location is selected.

Regularized Evolution mutates a uniformly distributed random layer in the architecture sequence.
Because the distribution is random in expectation, the model architecture sequence is mutated in the middle of the architecture sequence.
This has implications for transfer learning: when transfer learning is performed, all of the layers after a mutation are  need to be retrained.
This limits the number of layers that potentially could be transferred in expectation because fewer than 50\% of the sequence is going to be transferred.
The fact that in expectation, mutations happen in the middle of the architecture sequence, is something that potentially can be manipulated by manipulating the probability distribution used to select where to perform the mutation.
More work is needed to explore the impacts of this kind of biasing in favor of transfer learning.

Next, Regularized Evolution mutates that randomly selected layer into a random value not equal to the current value.
Now, if the processes were fully random, one would expect that the empirical probability of observing a particular variable node choice would be equal for each variable node.
But, the process is not fully random -- it is guided during stage 2 by the selection of the highest performing model in the sample.
We can exploit this lack of complete randomness in stage 2 to effect better caching in this stage whereas our options are more limited in stage 1.

\subsubsection{Observations}

First, we can consider the expectation from the theory that the mutations will tend to occur in the middle of the architecture sequence.
The effects of this are clearly seen in the tries in Figure~\ref{fig:trie}.
The tries show the most common prefixes are dense for the first 3 layers and become much sparser as the trie continues to the right.

Next, we consider how the choice of mutation from a particular variable node.
In the figure, for example, we see variable node choice 2 occurs 42\% of the time disproportionately many times of the 5 possible choices for this variable node.
Here again, we see selection effects as the search favors high-performing nodes.
If we consider the remaining variable nodes in the architecture sequence, we would observe the same behavior for the remaining slots -- certain variable node choices are far more popular than others.

What can we observe from the differences between the two different training lengths?
As we can observe, the variable node choices differ substantially between short training and longer training even for the same model.
For example in the shorter training, variable node choice 2 dominates the first choice, and choice 0 comes in a distant second, while for the longer training, variable node choice 0 dominates and choice 2 comes in a distant second.
This is interesting in that some configurations may comparatively perform more poorly with a few iterations of training suggesting that there are possibly grave consequences for naive few-shot.
Transfer learning may offer a possible solution here to low-quality performance with few-shot learning.
In transfer learning, model weights are transferred from one model candidate to another, frozen while non-transfer layers are trained, and then later unfrozen and ``fine-tuned''.
This process of transferring weights essentially provides additional epochs of training for a particular set of model weights as they are incrementally improved over the course of model training with each transfer potentially giving both the benefits of smaller numbers of epochs while providing some of the same types of quality improvements.
We will consider this aspect further later in our study
\subsubsection{Takeaways}
Both theory and empirical study show that the random selection of both location and specific mutations has an impact on the kinds of models observed during NAS.
Future work can exploit this to design effective caching systems for NAS using transfer learning and to bias the layer selection to favor transfer learning.
We finally observe that the transfer learning may help compensate for the smaller epoch counts and the differences in the search process.

\subsection{How do model evolution patterns occur in a distributed context where information is incomplete?}\label{sec:distributed}

\subsubsection{Motivation and Method}
Next, we zoom in further and can consider how the search process occurs across nodes by considering the search patterns of the particular model tensors as they are accessed by particular worker nodes in a distributed search powered by transfer learning.  Studying what tensors each worker gets can inform the search process in two ways when using transfer learning: 1) when a model tensor is repeated between multiple workers within a given point in time, broadcasting can be used to reduce the bandwidth required to distribute them to the workers  2) since model architecture sequences are approximately known in advance due to the deterministic nature of pseudo-random number generators, the workers that receive particular nodes can be chosen in such a way as to minimize communication if a particular model tensor is common among them. 3) We can observe the temporal locality of particular model tensors recurring which can inform worker local caching on the nodes.

\begin{figure}
    \centering
    \includegraphics[width=\columnwidth]{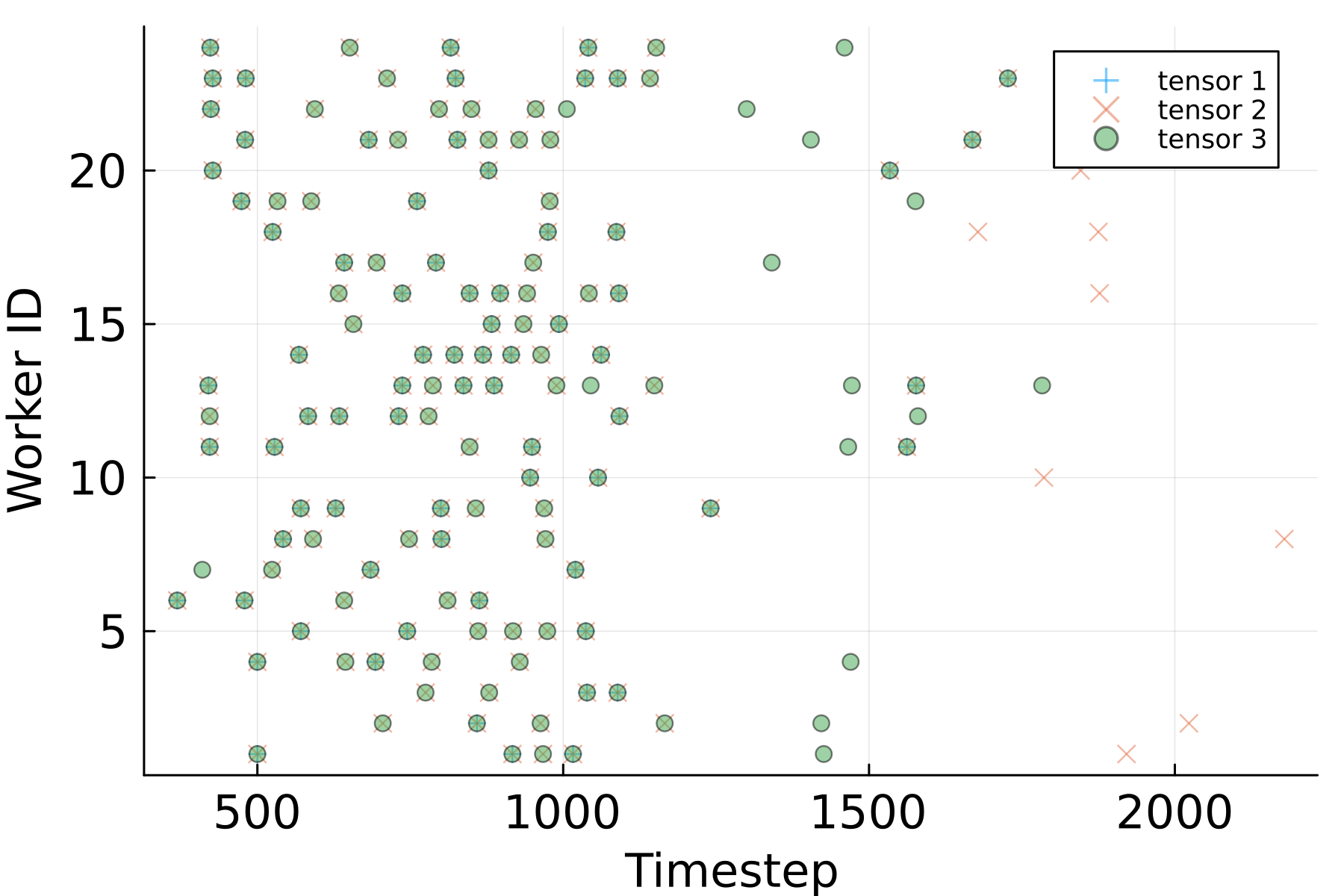}
    \caption{Which workers have which model layers transferred to them at which points in time.  This figure shows 2nd-4th most popular layers only for clarity,  The most frequently appearing model layer would cover this entire field.}
    \label{fig:workertransfer}
\end{figure}

We can use a scatter plot showing the worker ID, timestamp, and a few of the most popular tensors to study this aspect.
A scatter plot is useful here because it will allow us to see the locality of a particular tensor access as it is repeated both on a single client, and when multiple clients access a particular sequence across nodes simultaneously.
In Figure~\ref{fig:workertransfer} on the X-axis we show the timestamp of a model search beginning, on the Y-axis, we show the worker ID.  In the colors and markers, we show with transparency three symbols that represent the 2nd, 3rd, and 4th most popular model tensors.  From this, we can observe particular transfer patterns between nodes, and the co-occurrence of particular model tensors on various workers.  We omit the most popular model tensor for clarity as it occurs ubiquitously on all workers throughout the search process.

\subsubsection{Algorithmic Analysis}
Because regularized evolution is implemented on top of a distributed task pool, we know that accesses for particular tensors will be distributed across the system.
We further know based on the structural distribution we studied in Figure~\ref{fig:trie} that over the course of the search, access patterns for particular tensors between and within workers will recur especially in stage 2 of Regularized Evolution.
However, even with seeded random number generators, the variation in the times to train and evaluate models will introduce non-determinism into the NAS process because the different orders of model completion will result in differing populations that are sampled to determine which models to evolve.
Therefore, we will need to study this aspect empirically to draw conclusions about how sampling occurs between nodes.
The degree of non-determinism could potentially be decreased by waiting to assign workers to candidates until a certain number of workers have all completed their tasks.
Since candidate generation through mutation is fast relative to candidate evaluation through model training, this should not cause task starvation beyond the delay introduced by waiting for a group of tasks to finish.
This has the trade-off of increased determinism and information with the additional wait caused by the delay.
We can upper-bound this trade-off by using the formula for the average $k^{th}$ order statistic drawn \cite{royston_algorithm_1982} from a normal distribution and with appropriate scaling constant \cite{elfving_asymptotical_1947}.
We can upper bound the size of this delay on average by $E(s,w) - E(w,w)$ where $E(r,w) = \Phi^{-1}(\frac{r-\pi/8}{w-\pi/4+1})$ where $\Phi^{-1}$ is the inverse cumulative distribution function of normal distribution of the training times, $s$ is the number of workers that we wait to complete before assigning new candidates to workers, and $w$ is the total number of workers such that $s \leq w$.
We know this is an upper bound because it models the worst case because it models the case where all workers begin at the same time.  On subsequent iterations, there would be workers that did not finish when we last assigned candidates to workers but would now complete sooner in the next decision relative to other workers that started at the beginning of the new quanta.

\subsubsection{Observations}
We can observe two major trends from Figure~\ref{fig:workertransfer}.
First, there are clusters of particular model layers across several workers that are temporally located.  For example, workers 11-13, 20, and 23-25 access tensor 3 at nearly the same time. This temporal clustering could inform how information about these layers is broadcasted for example tree-based broadcasting, or job co-location to place related jobs on nearby nodes where they more easily share information.
Additionally, we can observe that a small scheduling quanta could be introduced with minimal effect on the scheduling time due to the tight verticle clustering seen between workers.

Second, we can observe that a particular layer can be frequently accessed on a particular worker ID.  For example, worker 19 accesses tensors 1, 2, and 3 in repeated succession.
We can further observe that several workers transfer similar groups of tensors throughout the search as can be observed from the overlapping of the symbols in the scatter plot.
This motivates the use of local caching on the workers of the most popular nodes, and potentially even grouping of commonly co-accessed tensors to perform more efficient bulk data transfers.

These two pairs of findings confirm our intuition that  there will be a locality of tensor accesses both between workers and within a worker.

\subsubsection{Takeaways}
There are temporal localities of a particular tensor both on a particular worker and between multiple workers.
The former motivates the use of worker-local caching of model layers, and the latter motivates the use of broadcasting and worker-to-task mappings that group similar workers together in the network.
We can model the upper bound of the  delay introduced by the scheduling quanta required to do this after the initial dispatch of the workers using a formula of the quanta size and the number of workers.

\subsection{What can be known about model quality during NAS over time?}\label{sec:time}
\subsubsection{Motivation and Method}
In some senses, this is both the foundational and most well-known aspect of NAS.
It is foundational in that it motivates the use is NAS despite its high computational cost -- higher quality models are worth some effort to develop them.
It is well-known that many studies of NAS consider the quality of the models over time \cite{real_regularized_2019}.
We include a brief treatment of quality during NAS here to motivate the broader study, but also to point out a key aspect that is hidden by the common formats of figures that describe the quality of a NAS and to highlight a key aspect of search results that will be important as we consider later research questions.

\begin{figure}
    \centering
    \includegraphics[width=\columnwidth,trim={0 0 0 1.5cm},clip]{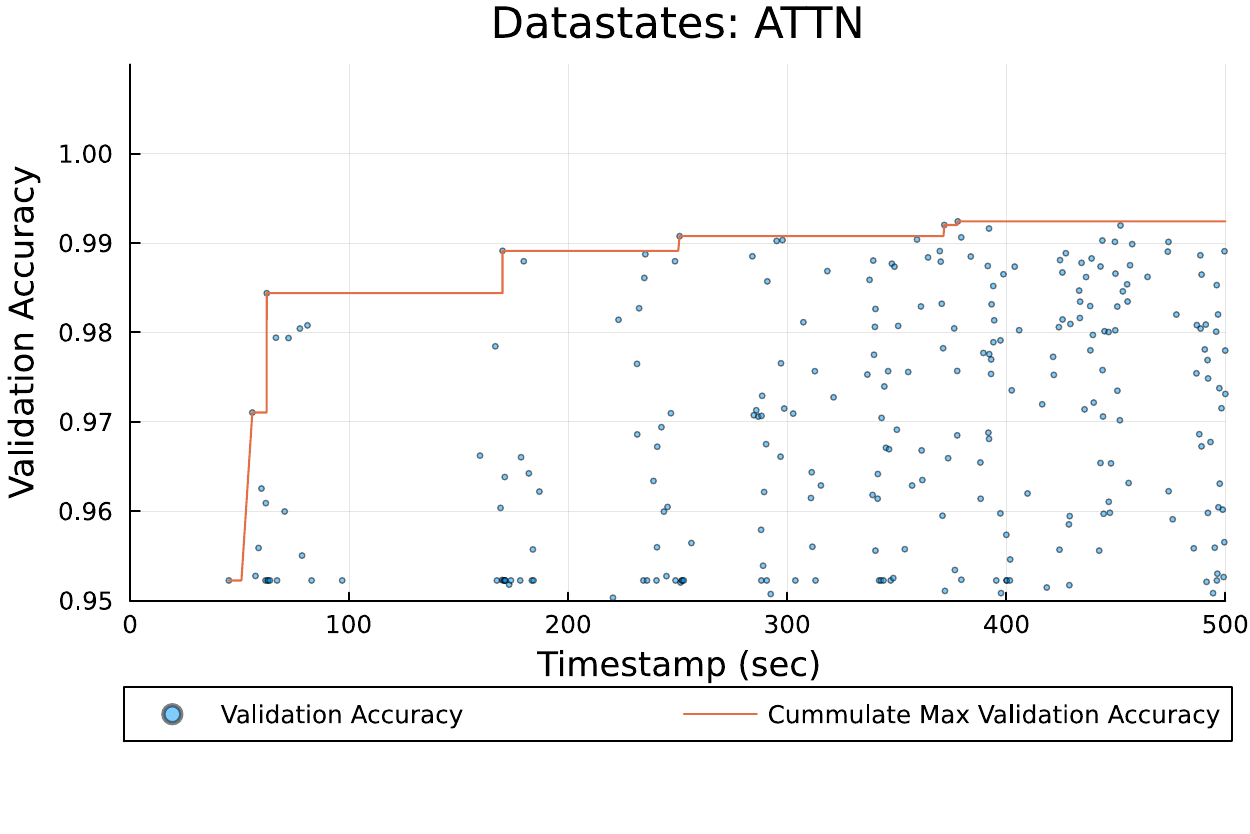}
    \caption{Cumulative and observed validation accuracy for Candle-ATTN over the course of a search with 1000 candidates without transfer learning.  Y-axis is trimmed to focus on high-quality candidates.}
    \label{fig:cumacc}
\end{figure}

To study model quality over time, we plotted a scatter plot of the quality of the model architectures against the end time of the competition of the search in Figure~\ref{fig:cumacc}.
We also include a line that shows the cumulative maximum accuracy of the models.
Together this combination clearly shows both the top performers, but also does not hide the presence of low performers throughout the search.
A band showing the performance of the top k models would have served the same purpose as displaying just the cumulative max, but we can make our point with just the top model performance.
This particular plot shows the cumulative accuracy for ATTN, but a similar plot is easily constructed for CIFAR-10.

We have trimmed the y-axis to make it easier to differentiate high-quality results from each other.  If we extended the y-axis to the minimum quality observed, there would be models with a validation score of less than .3 at all times during the search process.  This indicates that there are often missteps in the search processes which result in low-quality models.  This fact is often obscured by the formatting of figures in NAS papers which will either only show the max or top-1 accuracy as it is called in other papers, or by showing bands for the median, $90^{th}$, $95^{th}$, or $99^{th}$ percentile, or the interquartile range.
While it is true that high performers are important, and they are rightly the focus of prior papers, knowing of the presence of low-performers is important for the design of caching systems for example when performing transfer learning where caching them is wasted effort.

\subsubsection{Algorithmic Analysis}
First, let us consider the case of low performers.
Suppose that a new model has just been evaluated, and it has the worst accuracy in the current population, would it ever be transferred from?  Not in regularized evolution.  As such, you should skip I/O to store this model if it is only going to be used for transfer learning.  While this specific case of the worse performer is obvious, what about the next $s-2$ worse models? They too will also never be selected for transfer.  For higher quality models, we can calculate the probability that a model will be selected which is related to the hypergeometric distribution with a probability density function of  $p(X=k | X \sim H(N, K, n)) = \frac{\binom{K}{k} \binom{N-K}{n-k}}{\binom{N}{n}}$.  From this we can derive the probability that a model will be the basis for a transfer in a given timestep if $\lambda$ is the rank of the just-evaluated model, there are $P-\lambda$ models that it would overtake if they were sampled  is upper bounded by $p(X=0 | X \sim H(P, P-\lambda, s))$ allowing us to estimate the probability of transfer \footnote{the full probability would account for the that the model with rank $\lambda$ was sampled, not merely that no models were selected with higher quality than it, but this simpler form requires less computation, is conceptually simpler, and is likely good enough for a fast caching decision. Determining the exact probability requires careful use of both Bayes theorem and the hypergeometric probability distribution}.  If this probability is suitably small, caching could be skipped.

Next, let us consider the case of high performers.
Improvements in maximum quality are often stepwise.  
This is intuitive.
Suppose we knew the quality of models in the search space,  each time we observe a new maximum performance, there are fewer and fewer models that remain in the search space that are both improvements over the current best-observed model, and unobserved themselves.
Additionally, we expect that over time the magnitude of performance increases would also tend to decrease.
This too is intuitive.
If the distribution of model performance is approximately Normally distributed as it is in \cite{naslib-2020}, we would observe that the probability of getting a result with a large increase in quality is subsumed by getting a mere incremental improvement in quality.
The implication of this for optimizing NAS is that there are diminishing returns the greater the longer the search proceeds.

\subsubsection{Obervations}
Now let us turn our attention to the actual results from Candle-ATTN by examining the low and high-quality performers
While the lowest performers are omitted from Figure~\ref{fig:cumacc}, we do in fact have low performers throughout the NAS process.
As stated in our analysis, these low performers are unlikely to ever be transferred and are confirmed in our traces.
Next, we turn our attention to high performers.
In this search, we observe improvements in the max quality occur around timestep 50, again around 175, 250, and 375.
In this case, Stage 1 of the Algorithm~\ref{alg:regevo} completes near timestep 100.
We can observe that there are consistent improvements in the quality of results found in the search in phase 2.
We further observe that as the search continues the size of the improvement tends to decrease.
We will explore this aspect of stepwise improvement further in subsection~\ref{sec:popularity}.

\subsubsection{Takeaways}
We find key insights with respect to both low and high-quality performers that influence the design of transfer learning for NAS.
For low performers, we can upper bound the probability that a model will be transferred in the subsequent search process and thus define a threshold for model quality that would suggest that it would never be transferred allowing us to skip storing it.
Additionally, for high performers, there are diminishing returns to network architecture search.

\begin{figure*}
     \centering
     \begin{subfigure}[b]{0.3\textwidth}
         \centering
         \includegraphics[width=\textwidth]{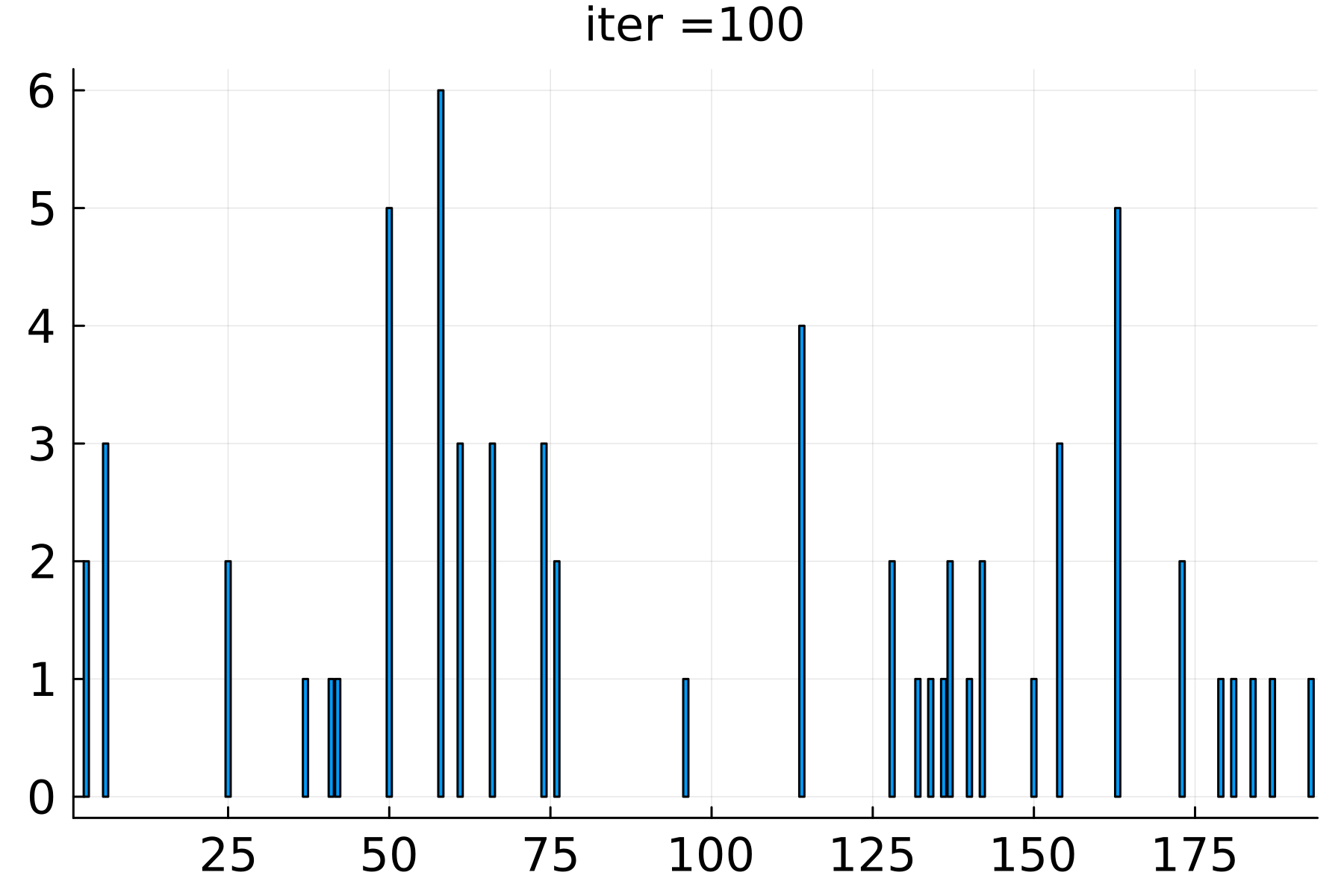}
         \caption{no transfer, candidate 100}
         \label{fig:notransfer100}
     \end{subfigure}
     \hfill
     \begin{subfigure}[b]{0.3\textwidth}
         \centering
         \includegraphics[width=\textwidth]{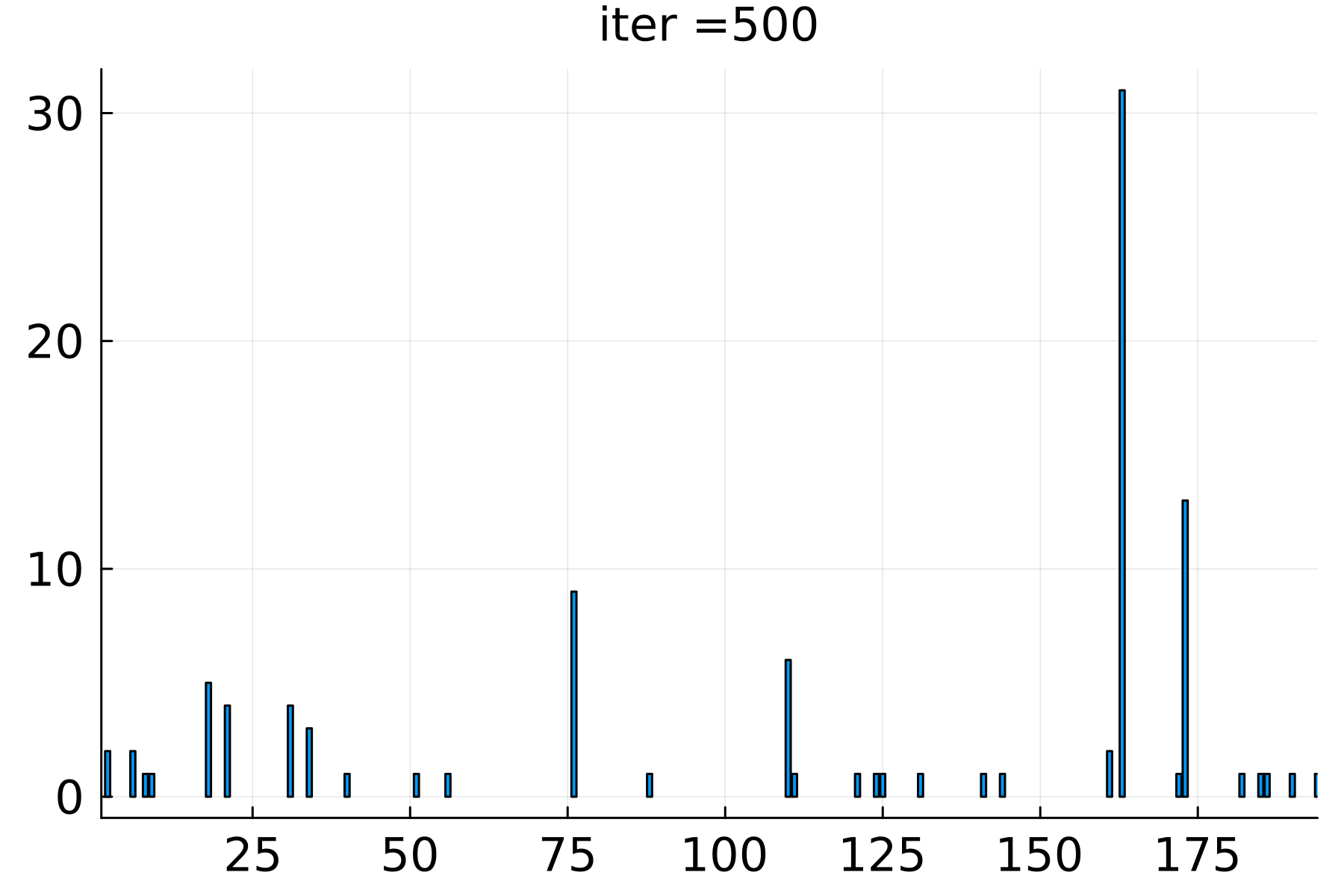}
         \caption{no transfer, candidate 500}
         \label{fig:notransfer500}
     \end{subfigure}
     \hfill
     \begin{subfigure}[b]{0.3\textwidth}
         \centering
         \includegraphics[width=\textwidth]{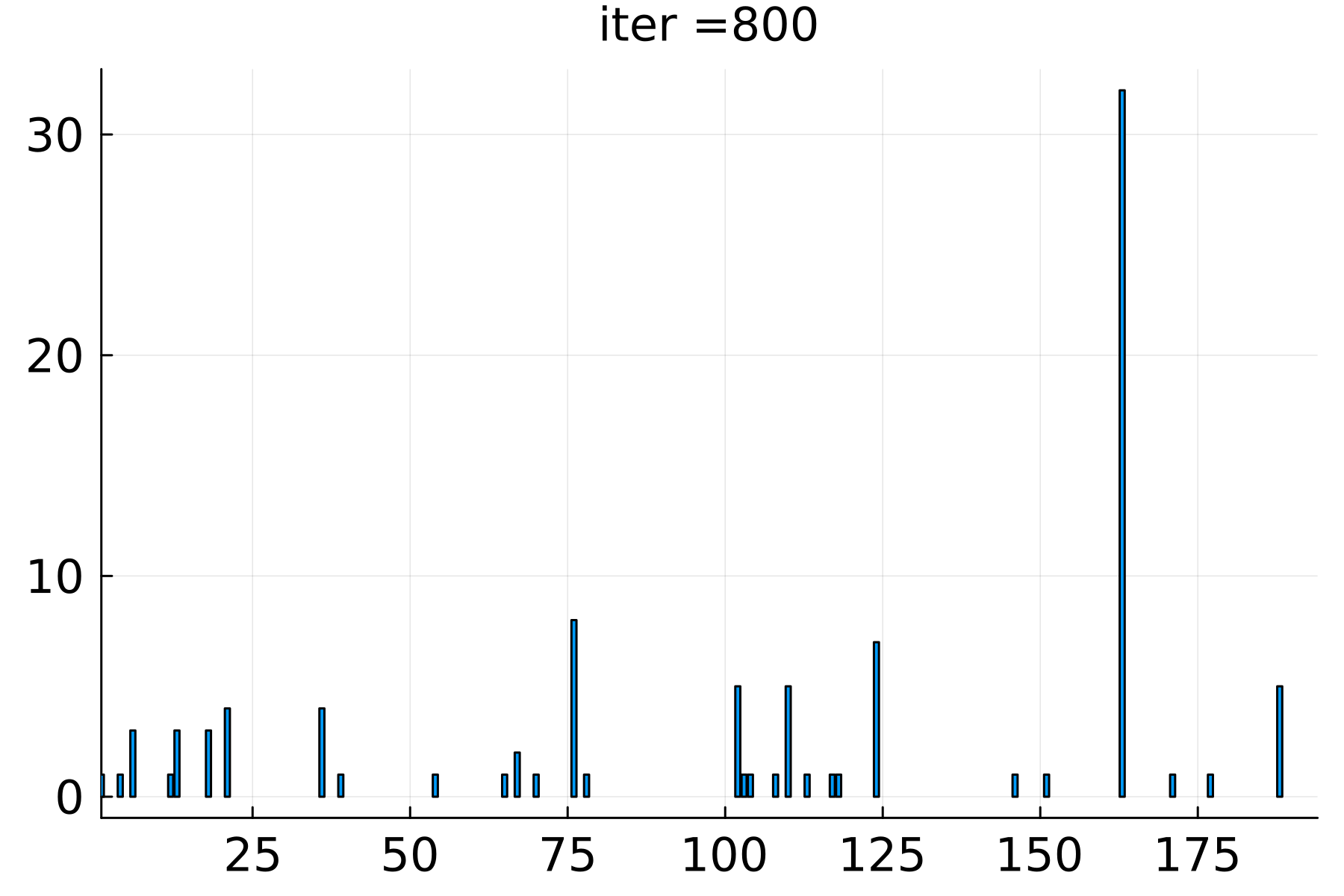}
         \caption{no transfer, candidate 800}
         \label{fig:notransfer800}
     \end{subfigure}\\
     \begin{subfigure}[b]{0.3\textwidth}
         \centering
         \includegraphics[width=\textwidth]{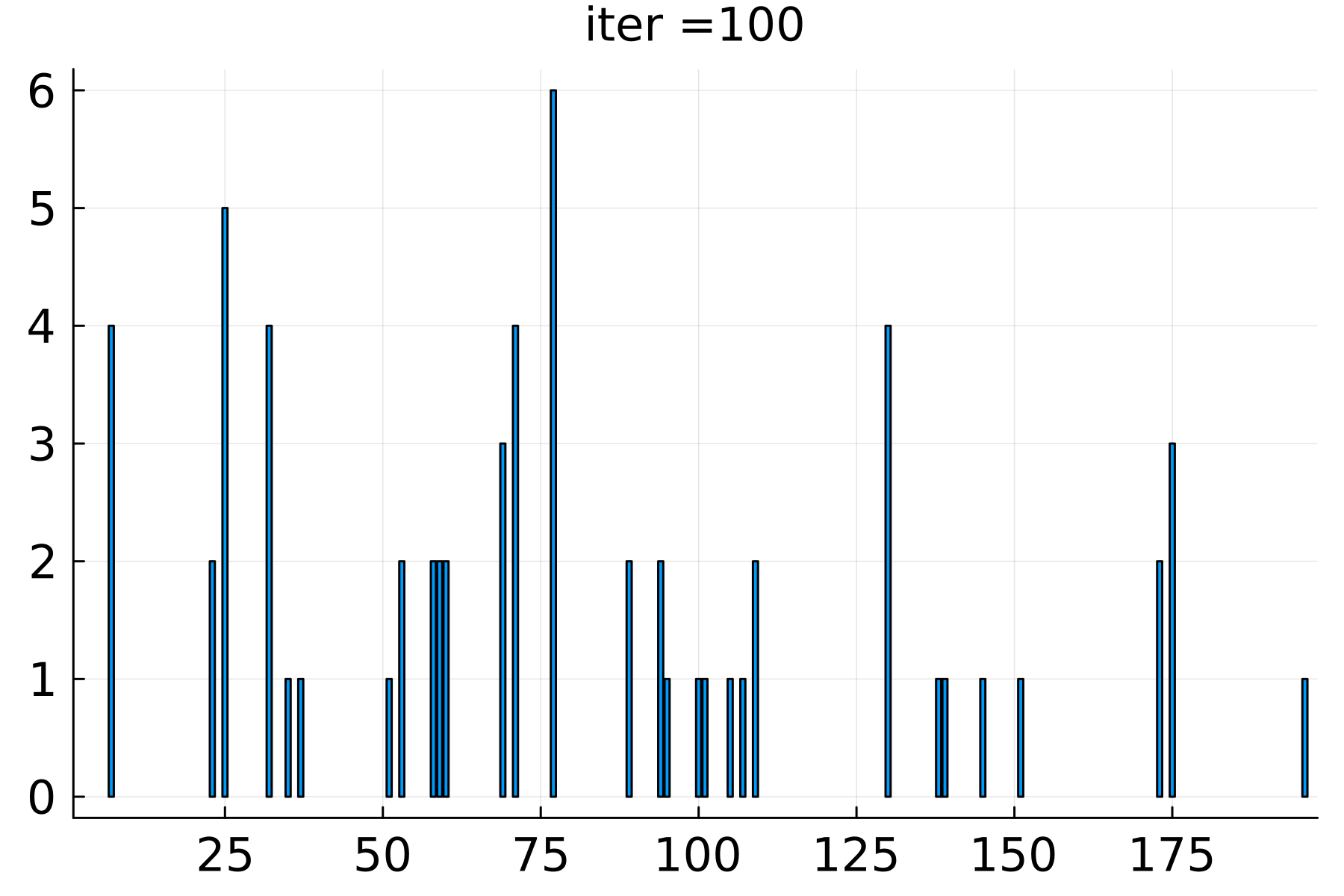}
         \caption{transfer, candidate 100}
         \label{fig:transfer100}
     \end{subfigure}
     \hfill
     \begin{subfigure}[b]{0.3\textwidth}
         \centering
         \includegraphics[width=\textwidth]{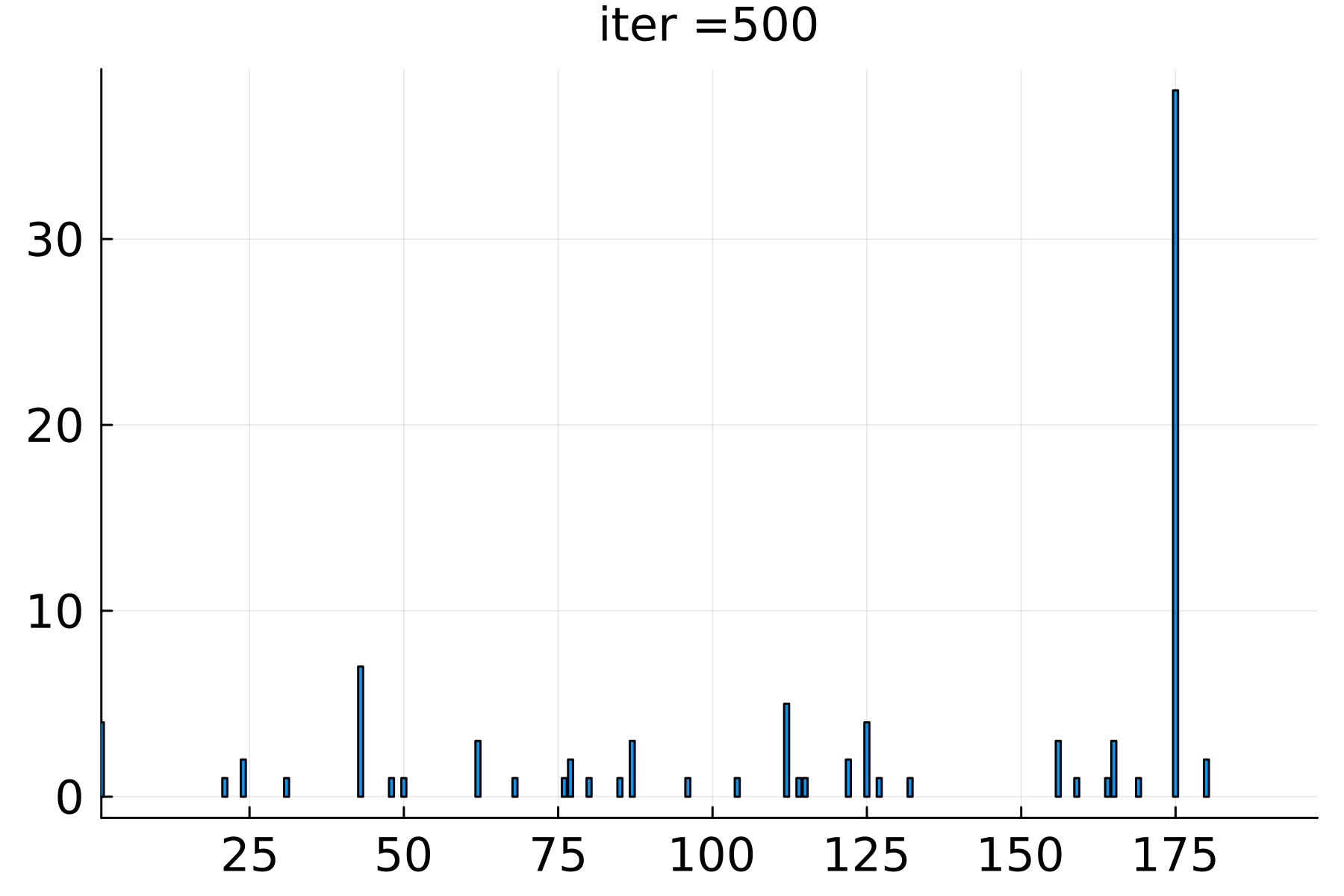}
         \caption{transfer, candidate 500}
         \label{fig:transfer500}
     \end{subfigure}
     \hfill
     \begin{subfigure}[b]{0.3\textwidth}
         \centering
         \includegraphics[width=\textwidth]{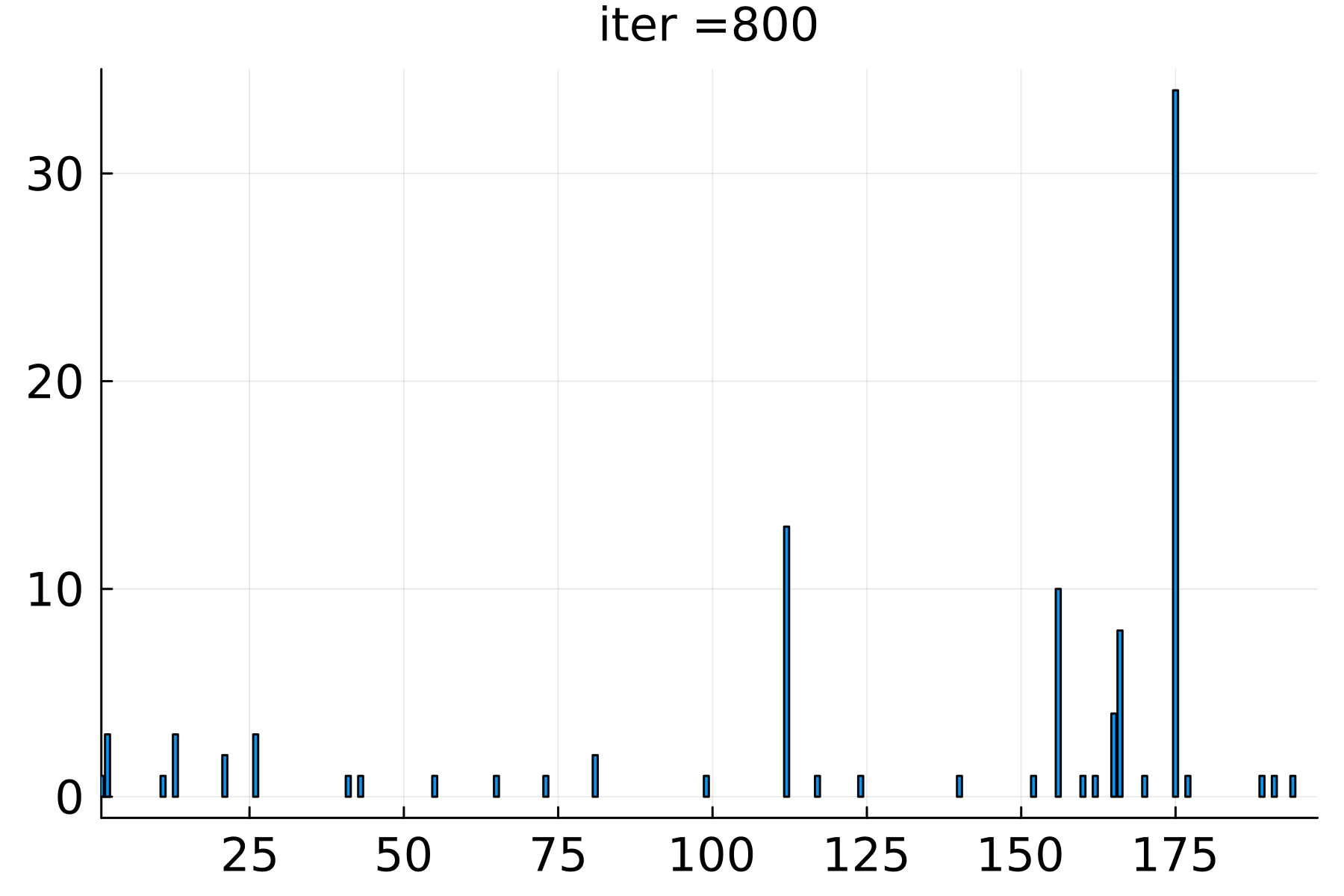}
         \caption{transfer, candidate 800}
         \label{fig:transfer800}
     \end{subfigure}\\
     \begin{subfigure}[b]{0.3\textwidth}
         \centering
         \includegraphics[width=\textwidth]{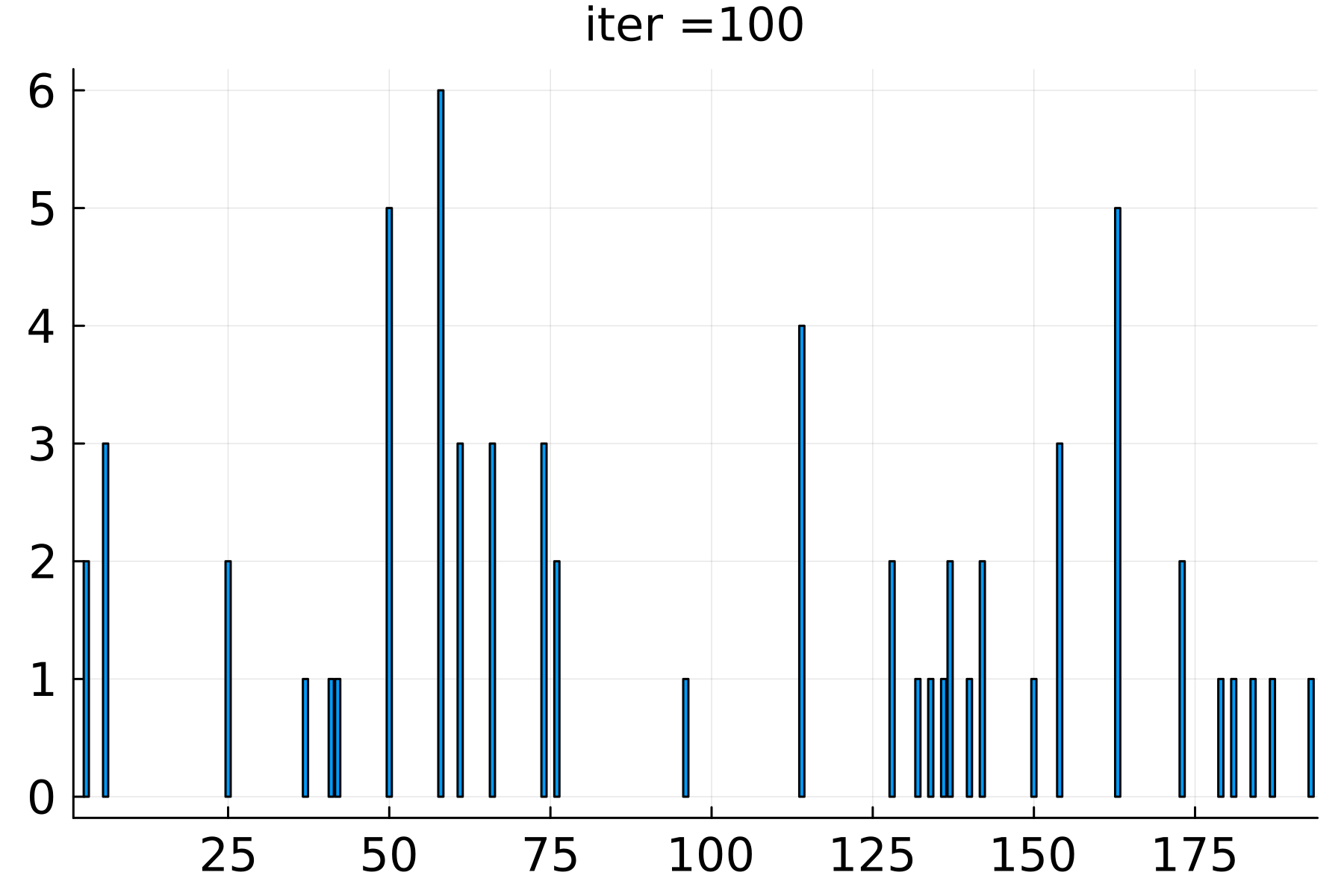}
         \caption{model 2, candidate 100}
         \label{fig:model2-100}
     \end{subfigure}
     \hfill
     \begin{subfigure}[b]{0.3\textwidth}
         \centering
         \includegraphics[width=\textwidth]{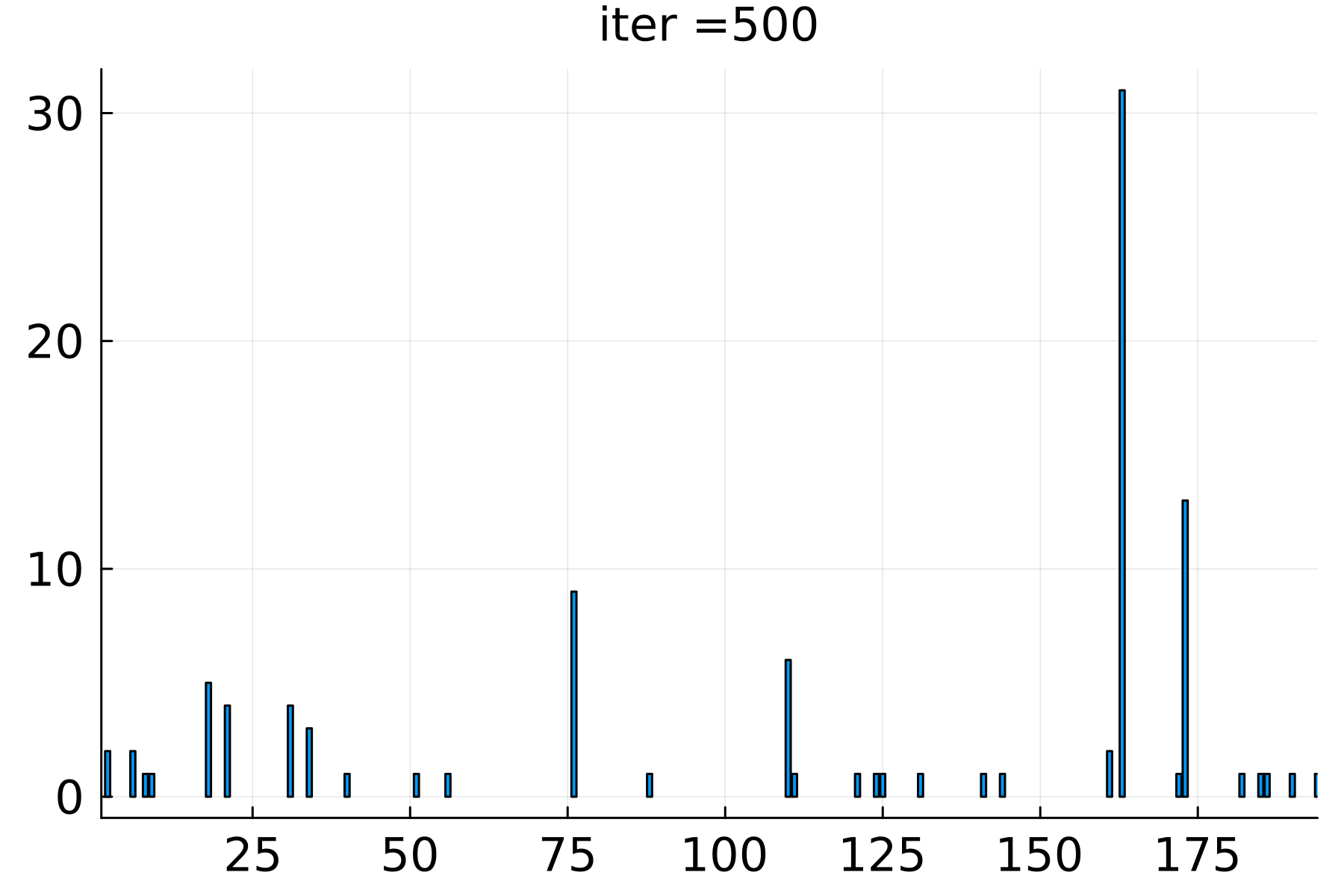}
         \caption{model 2, candidate 500}
         \label{fig:model2-500}
     \end{subfigure}
     \hfill
     \begin{subfigure}[b]{0.3\textwidth}
         \centering
         \includegraphics[width=\textwidth]{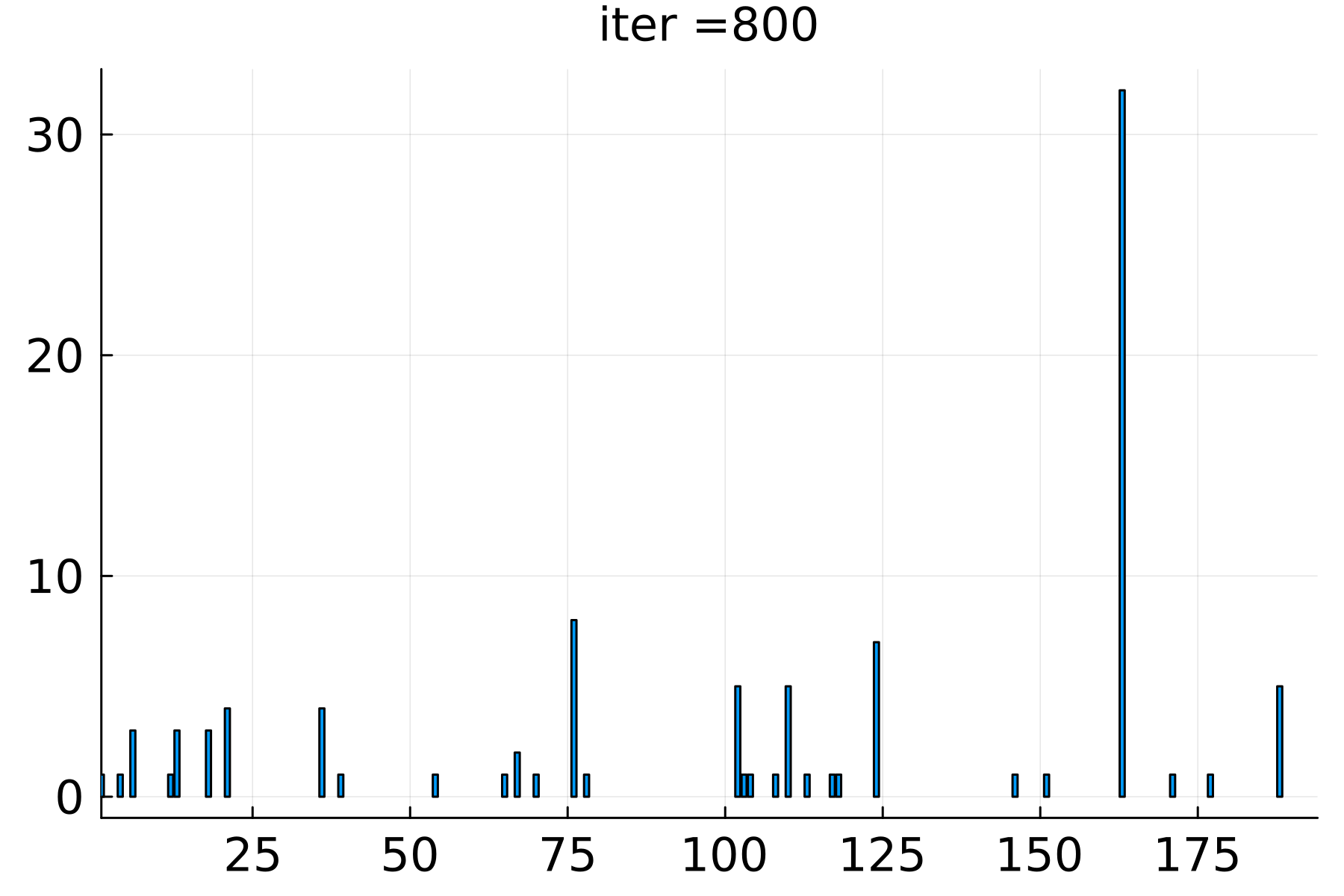}
         \caption{model 2, candidate 800}
         \label{fig:model2-800}
     \end{subfigure} 
        \caption{Evolution of the models over time with three transfer variants. Progressing from left to right shows three slices of time of the search process.  Each row shows a different model training variant.  The X-axis shows a unique id for a specified model prefix.  Prefix ids are consistent with all 9 subplots allowing comparisons of popular prefixes over time and search methods.  Y-axis shows the frequency of a particular model prefix in the last 100 model candidates considered (the current population)}
        \label{fig:evolution}
\end{figure*}

%Note: Heading can be made more consise such as 'What are the peaks and troughs of the model's access frequency?'
\subsection{When will a model architecture become popular and possibly subsequently become less popular and relevant?}\label{sec:popularity}
\subsubsection{Motivation and Method}
In this section, we combine aspects of our previous analysis to consider both the structure of transfers made, and how that evolves with the population changes over time.
Understanding how model structure evolves with the population has clear implications for the design of caching systems.
Beyond what we could observe from the structural analysis of Section~\ref{sec:structure} or Section~\ref{sec:distributed}, looking at how the population evolves over time gives us greater insight in to the exact prefixes of popular models that will occur.
Beyond what we could observe from the temporal analysis of Section~\ref{sec:time}, we can observe how changes in quality influence the choices of particular transfers of models which is ultimately the most important aspect of designing an efficient system for NAS.

To consider this aspect, in Figure~\ref{fig:evolution} we consider histograms of the architecture sequences from sliding windows of the traces over time corresponding with the population.
Histograms are useful in showing the distribution
By forcing the sliding window to coincide with the population, we can see how the populations' distribution of prefixes over the NAS.
We can only improve upon further this by utilizing animations that cannot be included due to the limitations of PDF to show the continuously evolving architectures that we will make available as supplementary material upon publication of this work\footnote{\url{https://www.zenodo.com/}}.
Since we cannot use full animations, we look at 3 distinct snapshots in time.

In each x-axis, we display a prefix id for each unique prefix of the architecture sequence of a given length (in this case 3) of models evaluated in the last population size number of models (in this case 100).
The prefix ids are common within a row of figures (e.g. id=50 on row 1 is the same prefix in all three columns), but are distinct between rows (e.g. id=50 on row 1 refers to a different prefix than on row 2), within a row, prefix ids are sorted based on the lexicographical ordering of their variable node choices and effectively arbitrary.
While it is difficult to tell, there are slightly different numbers of prefix ids in each row.
By making the prefix IDs consistent across the timesteps, we can see how different prefixes become more or less popular over time.
In each y-axis, we display the frequency that that prefix occurred in the last population size number of models.
Note that the y-axis ranges differ between subplots.
The columns represent a particular timestep of the search process, and because each search examines 1000 model candidates,  are the same across each model space considered so one can see the evolution of the search process over time.
Lastly, the first two rows are from CANDLE-ATTN with two different seeds, and the latter represents a run from CIFAR-10.

\subsubsection{Algorithmic Analysis}
Let's first consider how the maximum number of duplicates we expect to observe during stage 1 of Algorithm~\ref{alg:regevo}.
During stage 1, there is no direction to the search process.
A generalization of the birthday paradox \cite{diaconis_methods_1989} states that with probability $p$ that a particular prefix out of a total number of prefixes $c$ will occur at least $k$ times when the sample size is at least  $\left[c^{k-1} k! \log_e{\left(\frac{1}{1-p}\right)}\right]^{1/k}$.
However, we can also observe duplicates because of the state space itself -- not all possible prefixes are valid because they would result in an invalid model, and thus not prefixes of a given length do not have uniform probability.

Next, we can relate the time between when an improvement in the maximum quality is observed to when we expect to see changes in the distributions of model prefixes.
Models are selected for mutation whenever they are 1) first selected randomly from the population, and then 2) subsequently the highest performing model from the sampled subset.  Once a model sets a new cumulative maximum accuracy, it will then dominate any set of models that it is then sampled in making it a like source as a prefix.
We can compute the expected number of samples until this new maximum is selected and then mutated using the expected value of a geometric distribution.
This delay from when an improvement is found to when the improvement is then itself improved upon explains the gaps  between the changes in the relative popularity of the model prefixes we observe with the highest quality.

\subsubsection{Observations}

In stage 1, we observe for each of the three configurations that we have no model prefixes dominating the population.
The most commonly occurring prefixes constitute 6 out of the last 100 candidates.
This value while high, is still consistent with the theory regarding the generalized birthday paradox.
Beyond this effect, a specific concentration of the most popular prefixes can be attributed to the few cases of model prefixes that are slightly more popular than others due to the invalidities of some configurations.

Next, we can consider the next two columns which are reasonable distances into stage 2 of Algorithm~\ref{alg:regevo}.
In this column, we observe a few models with many more visits than others -- several near or exceeding 30 visits.
At this level of popularity, we cannot attribute the popularity of a particular prefix id to the generalized birthday paradox or the state space's subtle biases for particular prefix sequences.
Instead, we have a strong intuition that the regularized evolution algorithm is heavily selecting common prefixes.
We can see a first ``tier'' of prefix that appears in 30\% or more of the last considered models.
This first tier of model once it becomes popular, tends to dominate the search for the remaining duration and is seldomly dethroned.
We can also observe that there is a second ``tier'' of model prefixes that have more than 1 or 2 instances but also do not have more than 25 instances.
It is more common that these second ``tier'' peaks will move over time and other model alternatives drive performance.
A good example of this can be seen between \ref{fig:transfer500} and~\ref{fig:transfer800} where two tiers of two peaks appear between prefix ids 150 and the dominant peak of 175.
However,  very few of the popular models coming out of Stage 1 remain so in Stage 2.
Lastly, there is a third tier of model prefix that seldom gets more than 3 transfers.
These very seldom get multiple instances.

Lastly, we can look at how things can differ across models and transfer and no-transfer cases by comparing the rows of the model.
While we cannot draw inferences from the model IDs between these models, we can observe individual models may have a slightly greater or fewer number of second-tier peaks, and their relative size may fluctuate over the candidates, the same relative structure persists with one or at most 2 dominant peaks and a smaller number of second-tier peaks, and many entries with only 3 or fewer transfers.

\subsubsection{Takeaways}

What we describe as a three-tier nature of popularity has implications for the design of caching systems when using transfer learning in the context of NAS.
Specifically, in addition to not caching models in the lower levels of quality, it is likely not worth widely caching models until they have more than 5 transfers in the last 100.
The empirical probability that a model is a donor for mutation is substantially higher given that it has been a donor at least 5 times in the past.
Additionally, the gap between when a model is improved, and when it begins being used for transfer implies that in expectation, there is a timing opportunity to provide time to prefetch the model to the clients.

\section{Conclusions and Future Work}

This work developed a methodology and then performed a characterization study to study how models are produced and evaluated by regularized evolution in network architecture search.
We answered four key research questions regarding the structure of model candidates selected by this algorithm, how evolution patterns change in the context of NAS powered by asynchronous distributed workers where there is incomplete knowledge of model performance, and how model quality evolves over time,  and finally how portions of models become popular and subsequently become unpopular.

The answers to these questions set a path towards improving the scalability and performance of network archtiecture search using regularized evolution and other genetic algorithms.
There are three clear directions for future work:
1) I/O and caching for transfer learning -- our work proposed caching heuristics to be used based on the popularity of model layers.  Future work should evaluate these proposed heuristics in a model repository such as \cite{madhyasthaDStoreLightweightScalable2023} to alleviate I/O bottlenecks from using transfer learning with network architecture search.
2) Improvements to NAS scheduling -- our work identified some of the trade-offs along the continuum of batch scheduling and continuous scheduling which trade delays for improved accuracy and determinism.  Future work should evaluate these trade-offs in the context of a complete system where the increased quality could be more directly compared to the increased runtime.
3) Improvements to genetic search algorithms for NAS -- our work identified that using NAS with transfer learning is hampered by the limited expected number of layers transferred which could be addressed by weighting later nodes in the architecture more heavily for mutation.  Future work should evaluate the impact of these trade-offs on model quality.

\section*{Acknowledgments}
This material is based upon work supported by the U.S. Department
of Energy (DOE), Office of Science, Office of Advanced Scientific
Computing Research, under Contract DE-AC02-06CH11357.

\bibliographystyle{IEEEtran}
\bibliography{conference_101719}

\begin{thebibliography}{10}
\providecommand{\url}[1]{#1}
\csname url@samestyle\endcsname
\providecommand{\newblock}{\relax}
\providecommand{\bibinfo}[2]{#2}
\providecommand{\BIBentrySTDinterwordspacing}{\spaceskip=0pt\relax}
\providecommand{\BIBentryALTinterwordstretchfactor}{4}
\providecommand{\BIBentryALTinterwordspacing}{\spaceskip=\fontdimen2\font plus
\BIBentryALTinterwordstretchfactor\fontdimen3\font minus
  \fontdimen4\font\relax}
\providecommand{\BIBforeignlanguage}[2]{{%
\expandafter\ifx\csname l@#1\endcsname\relax
\typeout{** WARNING: IEEEtran.bst: No hyphenation pattern has been}%
\typeout{** loaded for the language `#1'. Using the pattern for}%
\typeout{** the default language instead.}%
\else
\language=\csname l@#1\endcsname
\fi
#2}}
\providecommand{\BIBdecl}{\relax}
\BIBdecl

\bibitem{wozniak_candlesupervisor_2018}
\BIBentryALTinterwordspacing
J.~M. Wozniak, R.~Jain, P.~Balaprakash, J.~Ozik, N.~T. Collier, J.~Bauer,
  F.~Xia, T.~Brettin, R.~Stevens, J.~Mohd-Yusof, C.~G. Cardona, B.~V. Essen,
  and M.~Baughman, ``\BIBforeignlanguage{en}{{CANDLE}/{Supervisor}: a workflow
  framework for machine learning applied to cancer research},''
  \emph{\BIBforeignlanguage{en}{BMC Bioinformatics}}, vol.~19, no. S18, p. 491,
  Dec. 2018, number: S18. [Online]. Available:
  \url{https://bmcbioinformatics.biomedcentral.com/articles/10.1186/s12859-018-2508-4}
\BIBentrySTDinterwordspacing

\bibitem{nvidia_megatron-lm_2023}
\BIBentryALTinterwordspacing
{NVIDIA}, ``Megatron-{LM},'' Feb. 2023, original-date: 2019-03-21T16:15:52Z.
  [Online]. Available: \url{https://github.com/NVIDIA/Megatron-LM}
\BIBentrySTDinterwordspacing

\bibitem{egele_agebo-tabular_2021}
\BIBentryALTinterwordspacing
R.~Égelé, P.~Balaprakash, I.~Guyon, V.~Vishwanath, F.~Xia, R.~Stevens, and
  Z.~Liu, ``\BIBforeignlanguage{en}{{AgEBO}-tabular: joint neural architecture
  and hyperparameter search with autotuned data-parallel training for tabular
  data},'' in \emph{\BIBforeignlanguage{en}{Proceedings of the {International}
  {Conference} for {High} {Performance} {Computing}, {Networking}, {Storage}
  and {Analysis}}}.\hskip 1em plus 0.5em minus 0.4em\relax St. Louis Missouri:
  ACM, Nov. 2021, pp. 1--14. [Online]. Available:
  \url{https://dl.acm.org/doi/10.1145/3458817.3476203}
\BIBentrySTDinterwordspacing

\bibitem{balaprakash_autotuning_2018}
P.~Balaprakash, J.~Dongarra, T.~Gamblin, M.~Hall, J.~K. Hollingsworth,
  B.~Norris, and R.~Vuduc, ``Autotuning in {High}-{Performance} {Computing}
  {Applications},'' \emph{Proceedings of the IEEE}, vol. 106, no.~11, pp.
  2068--2083, Nov. 2018, number: 11 Conference Name: Proceedings of the IEEE.

\bibitem{real_regularized_2019}
\BIBentryALTinterwordspacing
E.~Real, A.~Aggarwal, Y.~Huang, and Q.~V. Le,
  ``\BIBforeignlanguage{en}{Regularized {Evolution} for {Image} {Classifier}
  {Architecture} {Search}},'' Feb. 2019, arXiv:1802.01548 [cs]. [Online].
  Available: \url{http://arxiv.org/abs/1802.01548}
\BIBentrySTDinterwordspacing

\bibitem{tan_survey_2018}
C.~Tan, F.~Sun, T.~Kong, W.~Zhang, C.~Yang, and C.~Liu,
  ``\BIBforeignlanguage{en}{A {Survey} on {Deep} {Transfer} {Learning}},'' in
  \emph{\BIBforeignlanguage{en}{Artificial {Neural} {Networks} and {Machine}
  {Learning} – {ICANN} 2018}}, ser. Lecture {Notes} in {Computer} {Science},
  V.~Kůrková, Y.~Manolopoulos, B.~Hammer, L.~Iliadis, and I.~Maglogiannis,
  Eds.\hskip 1em plus 0.5em minus 0.4em\relax Cham: Springer International
  Publishing, 2018, pp. 270--279.

\bibitem{liu_accelerating_2021}
H.~Liu, B.~Nicolae, S.~Di, F.~Cappello, and A.~Jog, ``Accelerating {DNN}
  {Architecture} {Search} at {Scale} {Using} {Selective} {Weight} {Transfer},''
  in \emph{2021 {IEEE} {International} {Conference} on {Cluster} {Computing}
  ({CLUSTER})}, Sep. 2021, pp. 82--93, iSSN: 2168-9253.

\bibitem{chollet_deep_2018}
F.~Chollet, \emph{Deep learning with {Python}}.\hskip 1em plus 0.5em minus
  0.4em\relax Shelter Island, New York: Manning Publications Co, 2018, oCLC:
  ocn982650571.

\bibitem{madhyasthaDStoreLightweightScalable2023}
M.~Madhyastha, R.~Underwood, R.~Burns, and B.~Nicolae, ``{{DStore}}: {{A
  Lightweight Scalable Learning Model Repository}} with {{Fine-Grain
  Tensor-Level Access}},'' in \emph{Proceedings of the 37th {{International
  Conference}} on {{Supercomputing}}}, ser. {{ICS}} '23.\hskip 1em plus 0.5em
  minus 0.4em\relax {New York, NY, USA}: {Association for Computing Machinery},
  Jun. 2023, pp. 133--143.

\bibitem{naslib-2020}
M.~Ruchte, A.~Zela, J.~Siems, J.~Grabocka, and F.~Hutter, ``Naslib: A modular
  and flexible neural architecture search library,''
  \url{https://github.com/automl/NASLib}, 2020.

\bibitem{balaprakash_DeepHyper_2018}
P.~Balaprakash, M.~Salim, T.~D. Uram, V.~Vishwanath, and S.~M. Wild,
  ``{DeepHyper}: {Asynchronous} {Hyperparameter} {Search} for {Deep} {Neural}
  {Networks},'' in \emph{2018 {IEEE} 25th {International} {Conference} on
  {High} {Performance} {Computing} ({HiPC})}, Dec. 2018, pp. 42--51, iSSN:
  2640-0316.

\bibitem{whitley_evaluating_1996}
\BIBentryALTinterwordspacing
D.~Whitley, S.~Rana, J.~Dzubera, and K.~E. Mathias,
  ``\BIBforeignlanguage{en}{Evaluating evolutionary algorithms},''
  \emph{\BIBforeignlanguage{en}{Artificial Intelligence}}, vol.~85, no. 1-2,
  pp. 245--276, Aug. 1996. [Online]. Available:
  \url{https://linkinghub.elsevier.com/retrieve/pii/0004370295001247}
\BIBentrySTDinterwordspacing

\bibitem{dong2020nasbench201}
X.~Dong and Y.~Yang, ``Nas-bench-201: Extending the scope of reproducible
  neural architecture search,'' 2020.

\bibitem{royston_algorithm_1982}
\BIBentryALTinterwordspacing
J.~P. Royston, ``\BIBforeignlanguage{en}{Algorithm {AS} 177: {Expected}
  {Normal} {Order} {Statistics} ({Exact} and {Approximate})},''
  \emph{\BIBforeignlanguage{en}{Applied Statistics}}, vol.~31, no.~2, p. 161,
  1982. [Online]. Available:
  \url{https://www.jstor.org/stable/10.2307/2347982?origin=crossref}
\BIBentrySTDinterwordspacing

\bibitem{elfving_asymptotical_1947}
\BIBentryALTinterwordspacing
G.~Elfving, ``\BIBforeignlanguage{en}{The {Asymptotical} {Distribution} of
  {Range} in {Samples} from a {Normal} {Population}},''
  \emph{\BIBforeignlanguage{en}{Biometrika}}, vol.~34, no. 1/2, pp. 111--119,
  Jan. 1947. [Online]. Available: \url{https://www.jstor.org/stable/2332515}
\BIBentrySTDinterwordspacing

\bibitem{diaconis_methods_1989}
P.~Diaconis and F.~Mosteller, ``\BIBforeignlanguage{en}{Methods for {Studying}
  {Coincidences}},'' \emph{\BIBforeignlanguage{en}{Journal of the American
  Statistical Association}}, 1989.

\end{thebibliography}
\end{document}